# Finding Shortest Path for Developed Cognitive Map Using Medial Axis

| Hazim A. Farhan | Hussein H. Owaied | Suhaib I. Al-Ghazi |
|---|---|---|
| Computer Science Dept. | Computer Science Dept. | Computer Science Dept. |
| Middle East University | Middle East University | Middle East University |
| Amman, Jordan. | Amman, Jordan. | Amman, Jordan. |
| hfarhan@meu.edu.jo | howaied@meu.edu.jo | suhaib.Ibraheem@yahoo.com |

*Abstract*— this paper presents an enhancement of the medial axis algorithm to be used for finding the optimal shortest path for developed cognitive map. The cognitive map has been developed, based on the architectural blueprint maps. The idea for using the medial-axis is to find main path central pixels; each center pixel represents the center distance between two side boarder pixels. The need for these pixels in the algorithm comes from the need of building a network of nodes for the path, where each node represents a turning in the real world (left, right, critical left, critical right…). The algorithm also ignores from finding the center pixels paths that are too small for intelligent robot navigation. The Idea of this algorithm is to find the possible shortest path between start and end points. The goal of this research is to extract a simple, robust representation of the shape of the cognitive map together with the optimal shortest path between start and end points. The intelligent robot will use this algorithm in order to decrease the time that is needed for sweeping the targeted building.

## I. INTRODUCTION

Human in many situations needs to have basic information about the place or building he will surface so that he can navigate inside it. Therefore Human start building knowledge about the spatial that he will face either acquiring that from a person who knows the place or using the information gained from reading the architectural blueprints of a building. However Schmidt in 2007 discuses that: "When animals (including humans) first explore a new environment, what they remember is fragmentary knowledge about the places visited. Yet, they have to use such fragmentary knowledge to find their way home. Humans naturally use more powerful heuristics while lower animals have shown to develop a variety of methods that tend to utilize two key pieces of information, namely distance and orientation information." [1].

An intelligent robot is a remarkably useful combination of a manipulator, sensors and controls. The computer and the robot have both been developed during recent times. The intelligent robot combines both technologies and requires a thorough understanding and knowledge of mechtronics [2]. Much of the research in robotics has been concerned with vision and tactile sensing. For example, one of the most important considerations in using a robot in a workplace is human safety [3]. A robot equipped with sensory devices that detects the presence of an obstacle or a human worker within its workspace and automatically stops its motion or shuts itself down in order to prevent any harm to itself for the human worker is an important current implementation in most robotics work cells.

Thrun in 2002 made a good survey discussing the techniques that have been used on intelligent robot mapping [4]. There are many concepts have raised, some of it researched concepts, that psychologists think how human thinks and build maps and addressed it with the machines architecture [5]. While others tried to find a new mapping concepts to perform human jobs far away from human thinking concepts [6]-[8]. Resent study about that was to build a network of camera systems linked with a centralized unit that give direction for a "Town Robot" by [9]. However, recent studies assure that to make intelligent robots perform human jobs we need to use the concept of human intelligence [10]. To reach the best performance for a robot to be intelligent, human intelligence concepts are now being studied to be implemented on intelligent robots [11] claim that "Object tracking is so basic and in high demand that it is an indispensable component in many applications including robot vision, video surveillance, object based compression, etc.". Also other researcher they study the control issues associated with the non-linear systems in real time using cost effective data acquisition system [12].

Most recent research on this field was made by [13], [14], trying to study human system in term of functions, since the framework model is for a Robot doing a job as human in a specified and specific domain.

Using architectural blueprint maps in building cognitive maps can facilitate even humans to better do their jobs, and to better build their cognitive maps. In reality a research on this field has discovered that even two years of work in the same building do not build a correct cognitive map like persons who used blueprints for the building from the first time. Moeser in 1988 discusses that: "The series of studies reported in his article examined the cognitive mapping systems of student nurses who had worked in the hospital for various periods of time [15]. After inspecting several different measures, it was concluded that the student nurses had failed to form 'survey'-type cognitive maps of the building even after traversing it for two years. A control experiment was tested, using naive subjects who were first asked to memorize floor plans of the building. These naive subjects performed significantly better on objective measures of cognitive mapping than did the nurses with two years' experience working at the hospital.

Intelligent robots presently needs to be more dynamic, reliable and having the ability to be mobilized even within indoor environments or buildings that is abandoned without the needs of guidance. This ability will facilitate the intelligent robot reaching its target in the simplest and shortest way [16].

The mentioned ability is more demanded for the intelligent robots with special jobs such as police man or fire fighters intelligent robots where human guidance is not possible. As known there are no satellite images for indoor environments, thus geographical maps are not useful inside such environments. Therefore chose of this work came on the architectural blueprints maps where the indoor environments are completely covered and a cognitive map have been developed based on blueprint map [17]. The developed cognitive map will be briefly described in next section before introducing the shortest path algorithm, since the algorithm based on the developed cognitive map.

## II. DEVELOPING COGNITIVE MAP

The methodology of creating the cognitive map using blueprint map as a knowledge-based system developed consists of many algorithms; the algorithms have been used by the inference engine of the knowledge based-system of the intelligent robot [18]. These algorithms are used for gaining knowledge from the architectural blueprint maps, to build the cognitive map. These algorithms respectively are:

1. Threshold algorithm.
2. Flood Fill Algorithm.
3. Start Point Finding.
4. End Point Finding.
5. Main Path Finding.
6. Targeted Object (Room) Finding.
7. Similarity Algorithm.

Fig.1 presents the output of this system, which consists of all the paths from start point to the end point, which is the target object (room). This map will be the input for the shortest path algorithms.

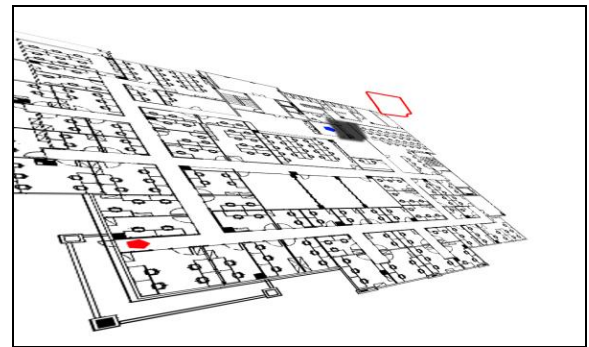

Fig.1: Main Path Extracted from the Blueprint

## III. SHORTEST PATH ALGORITHMS FOR THE DEVELOPED COGNITIVE MAP

The Optimal Shortest Path is the shortest path that has fewer turnings on it which means, less processing time during navigation. To find the optimal shortest path, the algorithms find the three shortest paths and select the path that has fewer nodes, and passes it on. The medial axis have been defined by Graf in 1999 which is "The medial-axis transformation is useful in thinning a polygon, or, as is sometimes said, finding its skeleton [19]. The goal is to extract a simple, robust representation of the shape of the polygon". In this research medial axis has been enhanced in order to be used for finding the shortest path. The following modules are used to find the optimal shortest path:

1. finding boarder pixels Module
2. Finding Central Pixels for Corridors Module
3. Finding Turnings Centers Module
4. Labeling the Medial-Axis Module
5. Finding the Shortest Path Module
6. Creating Graph (Connecting Nodes) Module
7. Floyd's Algorithm Module
8. The Optimal Shortest Path Module

### A. Finding Boarder Pixels Module

First it is important to group the mask main Path Mask with the two doors masks, doors Mask and targeted Door Mask. The new mask contains green shapes, which are doors from doors Mask and targeted Door Mask, and white shape, which is the main path as seen in Fig. 2.

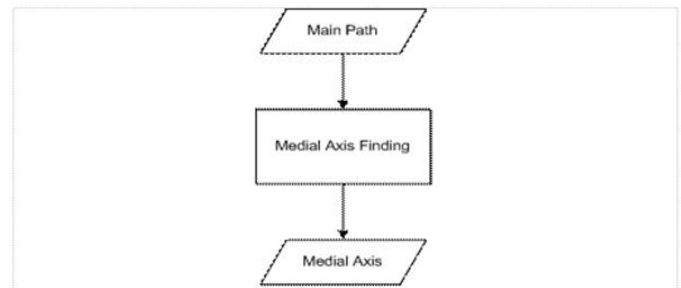

Fig. 2: Finding Boarder Pixels

The algorithm starts finding the boarder of the main path, where the boarder is any white pixel that at least one of its neighbors is not white pixel. All the boarder pixels are put on an array border Array. Algorithm 1 presents the Pseudo code finding boarder pixels.

```
//find boarder pixels and put them inside array
for each pixel P in the MainPath
if P eight neighbors == ffffff then continue
else
bourderArray[j] = P
j ++
end if
end for
```

Algorithm 1 Pseudo code finding boarder pixels

*B. Finding Central Pixels for Corridors*

There are two types of corridors for the main path, one of them is horizontal that includes typical horizontal and semi horizontal corridors, and the other one is vertical that includes typical vertical and semi vertical corridors.

The algorithm loops the grouped mask twice, one for finding center pixels for vertical corridors and the other loop for finding center pixels for horizontal corridors.

*1. Vertical Corridors*

Fig. 3 shows the path after finding center pixels for vertical corridors only using MATLAB R2009b.

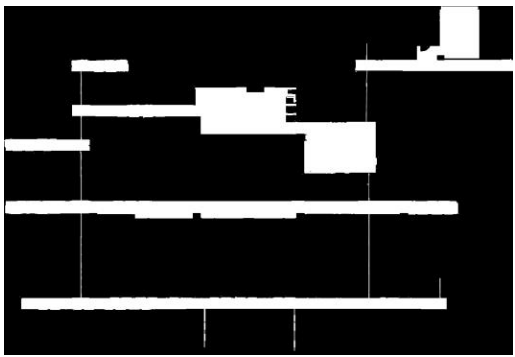

Fig. 3: the Path for Vertical Corridors Only

As shown in Algorithm 2, starts from the first boarder pixel in the mask, looping mask row by row, for the left boarder pixel (white boarder that its left neighbor pixel is nonwhite colored).

```
//for vertical and semi vertical paths
for each boarder pixel in boarderArray
//assure that it is a bottom boarder
if color(p4) !=ffffff then
for each p in boarderArray
if y(pbottom) == y(p) and x(pbottom)!=x(p) and color(pbottom)
!=ffffff and pbottom is not scanned and maxboarder>=(pbottom)-x(p)
>minboarder then
    pcenter = (y(p),(x(pnottom)-x(p))/2)
       pbottom scanned
  //checking for doors
  if any pixel between p and p(x(p)-20) ε doors then
   topDoors[i]=pcenter
  end if
  if any pixel between p and p(x(p)-20) ε targetDoor then
   targetTopDoors[i]=pcenter
  end if
  if any pixel between pbottom and p(y(pbottom)+20) ε doors then
   bottomDoors[i]=pcenter
  end if
  if any pixel between pbottom and p(y(pbottom)+20) ε targetdoor then
    targetBottomDoors[i] = pcenter
  end if
  for each pixel between p and pbottom including them
  if color(px) == ffffff then
   arraytoRemove[j]=px
   j++
  end if
 end for
 break the current loop
 end if
 if maxBoarder < x(pbottom) -x(p) then
 for each pixel between p and pbottomk including them
  bottomMaskShapr= px
 end for   end if  end for   end if   end for
```

Algorithm 2 Presents Pseudo Code for Vertical Corridors

The algorithm start searching for its right boarder, by increasing the x value of the left pixel until it reaches the right boarder (white boarder that its right neighbor pixel is nonwhite colored), if the distance between these two pixels is smaller than minBoarder value (the value that the corridor must be greater than) then all pixels between them including them are removed. And if the distance between these two pixels is greater than maxBoarder value (the value that the corridor must be smaller than) then these pixels are ignored for now. Otherwise the center pixel of these two side boarders is calculated and the Pcenter is then black colored, as equation 1.

Pcenter = (Y (P), X (Pright) - X (P) / 2) … equation 1

Where Pcenter means the pixel in the center, Y (P) means the value of Y coordinator of any Pixel, X (P) means the value of X coordinator of any Pixel, and the X (Pright) means the value of X coordinator of the pixel on the right of the center pixel. Then the algorithm puts the white pixels between the side boarders including the side boarders on an array named arraytoRemove, to delete this array later.

*2. Horizontal Corridors*

Same as vertical scanning, as shown in Algorithm 3, starting from the first boarder pixel in the mask, looping mask column by column, for the top boarder pixel (white boarder that its top neighbor pixel is nonwhite colored) the algorithm start searching for its bottom boarder, by increasing the y value of the top pixel until it reaches the bottom boarder (white boarder that its bottom neighbor pixel is nonwhite colored), if the distance between these two pixels is smaller than minBoarder value (the value that the corridor must be greater than) then all pixels between them including them are removed. And if the distance between these two pixels is greater than maxBoarder value (the value that the corridor must be smaller than) then these pixels are ignored for now. Otherwise the center pixel of these two side boarders is calculated and the Pcenter is then black colored, as equation 2.

Pcenter = (Y (P), X (Pbottom) - X (P) / 2) … equation 2

The algorithm puts the white pixels between the side boarders including the side boarders on an array named arraytoRemove, to delete this array later.

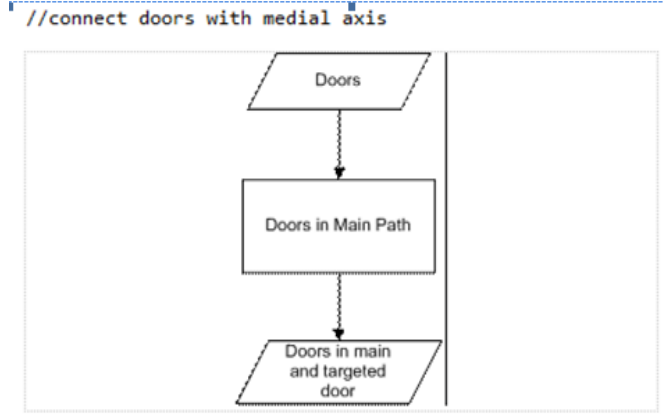

Fig. 4: Connecting Doors with Medial Axis Pseudo code.

Algorithm 4, searching 30 pixel on the top of P and 30 pixels on the bottom of Pbottom to find if there is a green pixel, which is the doors center pixels, if green pixel found, then the Pcenter is put on an array that collect all the doors on the main path and their coordinates doorsArray along with the "top" or "bottom" values, if the green pixel was within the targetDoorMask, then the variable endPointVar is set as Pcenter along with the "top" or "bottom" values.

```
    //finding central pixels between two side boarders
  //for horizontal and semi horizontal paths
  for each boarder pixel in boarderArray
  //assure that it is a left boarder
      if color(P4) != ffffff then for each P in boarderArray
      if y(Pright) == y(P) and x(Pright) != x(P) and color(Pright) != ffffff and
           Pright is not scanned and maxBoarder>= x(Pright)-x(P) >=minBoarder Then
              //checking for doors
              Pcenter = (y(P),(x(Pright)-x(P))/2)
              if any pixel between Pright and P(x(Pright)+20) ϵ doors then
                 rightDoors[i] = Pcenter
              end if
              if any pixel between Pright and P(x(Pright)+20) ϵ targetDoor then
                 targetRightDoors[i] = Pcenter
              end if
              if any pixel between P and P(x(P)-20) ϵ doors then
                 leftDoors[i] = Pcenter
              end if
              if any pixel between P and P(x(P)-20) ϵ targetDoor then
                 targetLeftDoors[i] = Pcenter
              end if
              Pright scanned
              for each pixel between P and Pright including them
              if color(Px) == ffffff then arraytoRemove[j] = Px
              end if
              end for
              break the current loop
      end if
          if maxBoarder< x(Pright)-x(P) then
             for each pixel between P and Pright including them
                rightMaskShape = Px
             end for
          end if
      end for
    end if
    end for
```

Algorithm 3: Center Pixels for Horizontal Corridors Pseudo code

```
//connect doors with medial axis
for each pixel in medial axis
   if x(P)+maxBoarder == dooronMain then
      if pixel(x(P)+maxBoarder) == dooronRoom then
         for each pixel from P to x(P)+maxBoarder
            Px = 000000
         end for
      end if
      remove dooronMain
      save pixel P, left to the short memory
   end if
   if x(P)-maxBoarder == dooronMain then
      if pixel(x(P)+maxBoarder) == dooronRoom then
         for each pixel from P to x(P)+maxBoarder
            Px = 000000
         end for
      end if
      remove dooronMain
      save pixel P, right to the short memory
   end if
   if y(P)+maxBoarder == dooronMain then
      if pixel(x(P)+maxBoarder) == dooronRoom then
         for each pixel from P to x(P)+maxBoarder
            Px = 000000
         end for
      end if
      remove dooronMain
      save pixel P, bottom to the short memory
   end if
   if y(P)maxBoarder == dooronMain then
      if pixel(x(P)+maxBoarder) == dooronRoom then
         for each pixel from P to x(P)+maxBoarder
            Px = 000000
         end for
      end if
      remove dooronMain
      save pixel P, top to the short memory
end if   end for
```

Algorithm 4: Connecting Doors with Medial Axis Pseudo code

As shown in Fig. 4 and algorithm 4 also searching 30 pixels on the left of P and 30 pixels on the right of P in order to find if there are green pixels; which are the door center pixels. If green pixels are found, then the Pcenter is put on an array that collect all the doors on the main path and their coordinates doorsArray along with the "left" or "right" values. But if the green pixels ware within the targetDoorMask, then the variable endPointVar is set as Pcenter along with the "left" or "right" values. Fig. 5, shows the path after finding center pixels for vertical and horizontal corridors. Fig. 6 presents the original Medial Axis after Applying the Algorithm on MATLAB R2009b

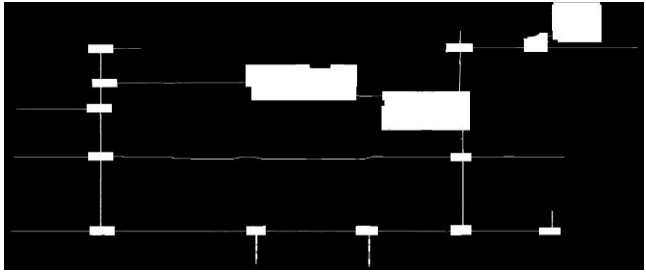

Fig. 5: Path after Finding Center Pixels for Vertical & Horizontal Corridors.

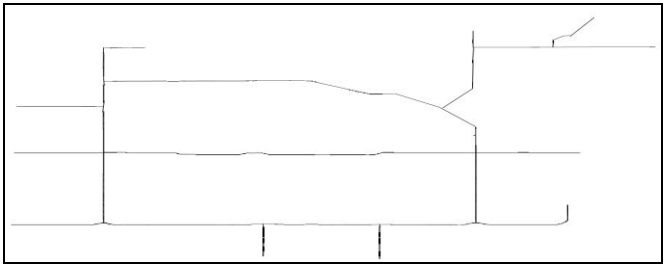

Fig. 6: Original Medial Axis after Applying the Algorithm on MATLAB R2009b.

C. *Finding Turnings Centers*

From applying the two above steps, almost all white pixels from the mask are removed, except white shapes that connected horizontal and vertical corridors, these shapes present the turning in the real world, and the nodes in the path graph. Since any pixel in these shapes cannot be either a center pixel or mid pixel (pixels between tow boarders), so these shapes where ignored by the conditions in the above two sections.

Using code in Algorithm-5, each turning shape is selected using floodfill8 function, then for each pixel in the shape if one of its neighbors (not all of them) is nonwhite colored pixel, then the pixel is a boarder pixel, it is removed. The algorithm repeats removing the turning boarder until finding a pixel that its eight neighbors' pixels colors are nonwhite, then this is a center pixel for the turning, that should be black colored and added to an array that contain all nodes of the main path named nodesArray.

The colors of these shapes are white, with at least two black pixels touching its boarder from more than one side, why? Because if there is one black pixel touching its boarder, then the corridor was not ignored by the conditions, (it is not a turning shape), and it has been already skeletonized in previous steps.

```
//turns from the previous step didn't included in its condition
//represent pixels that are in bottomMaskShape rightMaskShape
//and with black pixels touching its boarder from two sides at least
//repeat this step for all turnings
for all shapes
    for each borader pixel
        if one of P neighbors == 000000 then
            curFigNeig[i] = Pneig
            i ++
        end if
    end for
    for each pixel in the shape
        if P eight neighbors != ffffff then
            for each pixel from P to curFigNeig[i]
                Px = 000000
            end for
            turnings[j] = P
        else if one of P neighbors == ffffff then
            boarderPixel[i] = P
        end if
    end for
        remove boarderPixel array
    re-loop
end for

remove all pixels arraytoRemove
```

**Algorithm 5: Finding Turns Pseudo code**

The black pixels touching the turning shape are saved in an array for the current shape, before finding the center pixel for this shape. So when the center pixel is allocated the algorithm connect these pixels with the center pixel.

D. *Labeling the Medial-Axis*

Fig. 7 shows labeled medial axis with nodes on red for simulation purpose. As mentioned before, doors are already labeled in the medial-axis, also the end point variable was found, so the algorithm (Algorithm-6) searches for the start point in the medial-axis from the set of pixels in startPointMask, these values is put on a variable startPointVar.

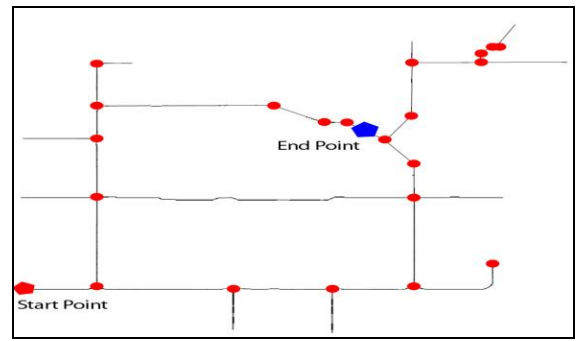

Fig. 7: Labeled Medial Axis with Nodes on Red for Simulation Purpose.

```
//Finding start point node in Map
For each pixel in MainPath
    If P ∈ startPointMask then
        startPointNode = P
        return
    end if
end for
```

Algorithm-6: Start Point Node in the Map

*E. Finding the Shortest Path*

The Idea of this algorithm is to find the possible shortest path between start and end points. In order to give the intelligent robot the required knowledge from the blueprints to sweep a building, the shortest path starting from StartPoint ending with EndPoint should be found in order to decrease the time that is needed for sweeping the targeted building. This Algorithm contains connected sub algorithms, which are described in next sections.

*F. Creating Graph (Connecting Nodes)*

As a result from the previous step, a collection of nodes have been extracted from the blueprint map. Each node must be connected to its neighbors, so that the nodes are in sets as a network, each node with its neighbor (i.e. (node1, node3), (node1, node4), (node2, node3) …).

Algorithm take the current pixel (initially it is the node itself), put it in a variable, then take one of its eight neighbors, if one of this pixel's eight neighbors is a node, then it finishes its job and return the node, otherwise, it takes the other neighbor from the eight neighbors of the pixel, if this pixel is not the same as the variable then it is set as the new variable, and checked for next neighbor again until next node is reached or a node with only one neighbor that is tagged in the variable, this means that a closed path reached. Two Cases are shown in Fig. 8.

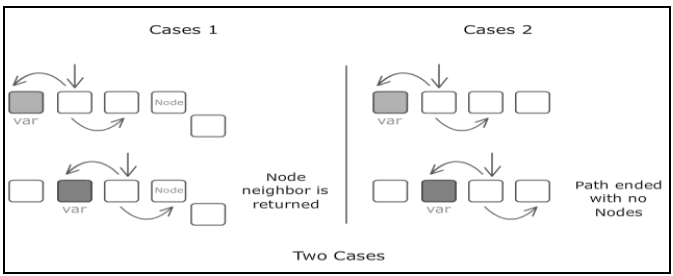

Fig. 8: Two Cases Appears on Connecting Nodes.

In order to find the neighbors, as shown in Fig. 9, each node boarder pixel labeled previously (including start and end points) is put in a function along with the node itself as a pixel that returns a node neighbor from one direction.

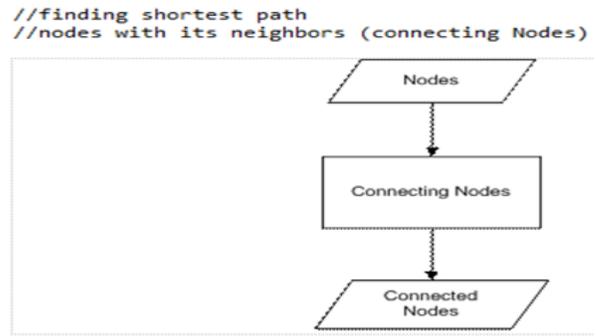

Fig. 9: Connecting Nodes.

In the case of finding the neighbor node, the distance between the two nodes is calculated and returned within the set (i.e. (node1, node3, 50), (node1, node4, 150), (node2, node3, 70) …), as seen in algorithm 7.

```
//finding shortest path
//nodes with its neighbors (connecting Nodes)
//For each node in the array of nodes find its neighbors
//of nodes and the distance between them
for each node
    for each node neighbor
        if color(neighbor) = 000000 then
            if getNextNode(node, neighbor) not in connectNodes[i] then
                connectNodes[i] = getNextNode(node, neighbor)
            end if
        end if
    end for
end for

//get next node function get the node and
//which node pixel neighbor number (1,2,3,…)
function getNextNode(node,neighbor)
    currentPixel = node
    nextPixel = neighbor(neighbor) where neighbor is not currentPixel
    while nextPixel is not node && nextPixel is not nothing
currentPixel = nextPixel
        nextPixel = neighbor(nextPixel) where neighbor is not currentPixel
    end while
 Return node & nextPixel
End function
```

Algorithm 7: Connecting Nodes Pseudo Code

*G. Floyd's Algorithm*

This algorithm has been previously declared as "It determines the shortest route between any two nodes in the network [20]. The algorithm represents an n-node network as a square matrix with n rows and n columns. Entry (~j) of the matrix gives the distance d ij from node i to node j, which is finite if the i is linked directly to j, and infinite otherwise". The algorithm finds the shortest path in a network of nodes. As it is build it takes the sets of nodes, where each node is put within its neighbor node along with the distance (weight) between the two nodes, also start and end points must be specified according to the flowchart in Fig. 10.

The algorithm 8 has been edited in this project to match on going algorithms, because the original one assumes directed paths, where no directed paths are considered in navigation.

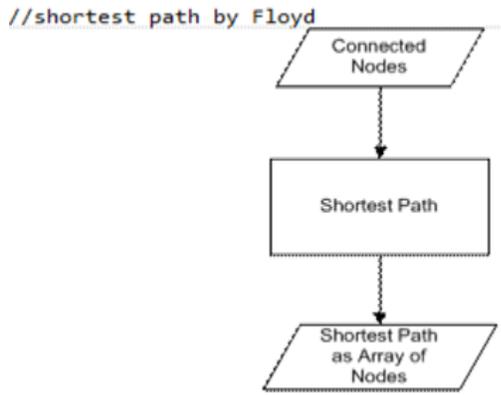

Fig. 10: Shortest Path by Floyd Algorithm

```
//shortest path by Floyd
start = startPoint
end = endPoint
d = connected nodes in a matrix
d{i in 1..n, j in 1..n}default M
rhs{i in 1..n}=if i=start then 1 else (if i=end then -1 else 0)
x{i in 1..n, j in 1..n}>=0
outFlow{i in 1..n}=sum{j in 1..n}x[i,j]
inFlow{j in 1..n}=sum{i in 1..n}x[j,j]
 minimize z: sum{i in 1..n, j in 1..n}d[i,j]*x[i,j]
subject to limit {I in 1..n}:outFlow[i]-inFlow[i]=rhs[i]
shortestPath[0] = startPoint
for i in 1..n-1
   for j in 2..n
      if x[i,j]=1 then
         shortestPath[counter]=(i,j)
         counter ++
      end if
   end for
end for
```

Algorithm 8: Shortest Path by Floyd Algorithm Pseudo code

*H. The Optimal Shortest Path*

The Optimal Shortest Path is the shortest path that has fewer turnings on it which means less processing time during navigation. To find the optimal shortest path, the algorithm finds the three shortest paths and selects the path that has fewer nodes, and passes it on.

*1. Translating to Directions*

Finding Shortest Path Algorithm returns the path as an array of nodes, this array needs to be translated as understandable directions to be built as a cognitive map.

If direction is found it is stored in the knowledge-base within the analyzed map knowledge section, and then next three nodes should be taken within the shortest path, tell reaching endpoint. Fig. 11 shows how angels are translated into directions.

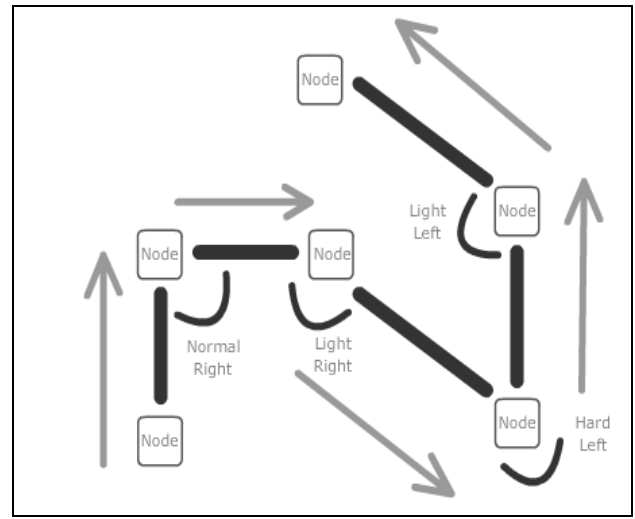

Fig. 11: Angels Translated into Directions.

As shown in Fig. 12, back to the labeled map, starting form StartPoint, the angle between each three nodes is calculated in order to find the direction three nodes presents, as specified previously in the knowledge-base, every angle presents a specific direction (i.e. 10-30: Hard right, 31-50: normal right, 180-200 right left and so on), by applying the algorithm 9.

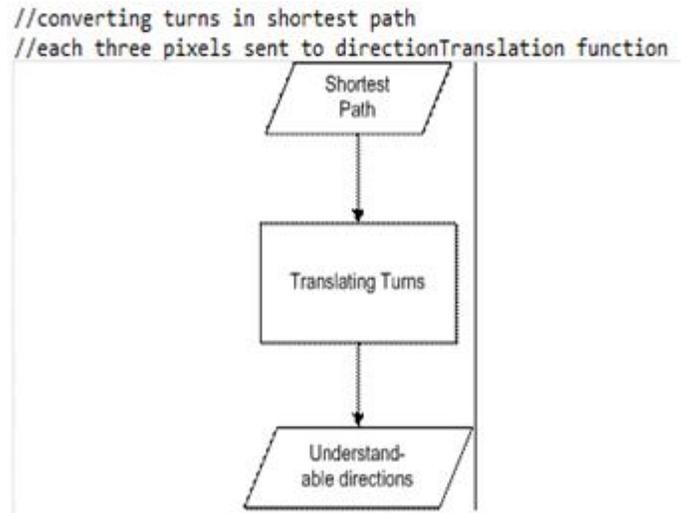

Fig. 12: Converting Nodes to Directions Algorithm Finding the Angle

In order to find the angle between each three nodes, the three nodes is presented in the Cartesian plane then the three line length could be found measuring the distance between each node from the another (the three lines are: one line from the first node to the second node, the second line is from the second node to the third node, and the third line is from the first node to the third node). These three lines present a triangle that its three lines are known, which means it can be solved (knowing other elements values which are the three angles of the triangle).

```
//converting turns in shortest path
//each three pixels sent to directionTranslation function
shortestPath[0]
while shortestPath[i] is not endpoint
   directions[i] = directionTranslation(shortestPath[i-1],
 shortestPath[i], shortestPath[i+1])
    i ++
end while
store directions to knowledge base

// directionTranslation function
function directionTranslation (pixel1, pixel2, pixel3)
   line1 = distance between pixel1 and pixel2
   line2 = distance between pixel2 and pixel3
   line3 = distance between pixel1 and pixel3
   if 45 >= calculateAngle(line1, line2, line3) >= 1 then
         return hard right
      end if
   if 150 >= calculateAngle(line1, line2, line3) >= 45 then
         return normal right
      end if
   if 180 > calculateAngle(line1, line2, line3) >= 150 then
         return light right
      end if
   if 210 >= calculateAngle(line1, line2, line3) >= 150 then
         return light left
      end if
   if 315 >= calculateAngle(line1, line2, line3) >= 210 then
         return normal left
      end if
   if 359 >= calculateAngle(line1, line2, line3) >= 315 then
         return hard left
      end if
end function
```

Algorithm 9: Converting Nodes to Directions Algorithm Pseudo code

The important angle is angle number one which present the angle or the complementary angle that is translated to direction.

At the end of this algorithm a knowledge base category is stored for this architectural blue print map, but a little knowledge still missing, this knowledge is in the regards of knowing which door is meant to be the targeted door (endpoint), whereas more than one door could be located on the mainPath, especially in the corridor that the targeted door exists, for example, three doors in the left and four doors on the right could exist, which will be somewhat confusing for even the human to find the targeted door from these doors, so the following algorithm was created to solve this issue.

2. *Allocate the Object (Door) in Directions*

As doors were previously allocated in the mainPath medial axis, it was not as much accurate to be used in this stage. Where it was allocated considering the optimal sweeping (sweeping from right to left or from down to top). In order to solve this issue, according to the flowchart in Fig.13, new doors positions should be applied the algorithm 10.

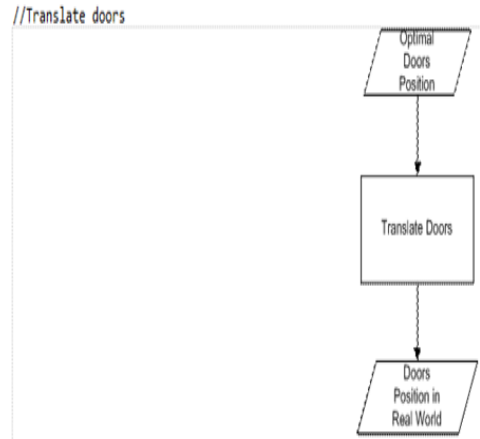

Fig. 13: Translating Doors Algorithm

As the corridor that the targeted door exists is the most important corridor in this stage, the algorithm 10 takes it (this corridor is presented as the node that comes before endpoint in the shortest path array). Applying this fact on the algorithm this corridor (node to endpoint) is taking, if the position of the node is on the right of the endpoint (in case of horizontal corridor), or on top of the endpoint (in the case of vertical corridor) then door positions is inverted (if left door then it is right door and if right door then it is a left door). Otherwise, door position remains as its default value (if left door then it is left door and if right door then it is right door).

Fig. 14 declares that the algorithm counts doors that come in the same side of the targeted door and locate the target door position (endpoint) and place this knowledge in the database.

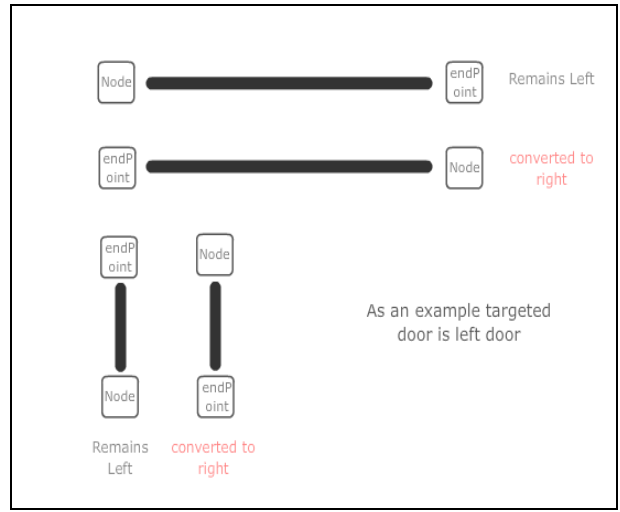

Fig. 14: Door Position Translation.

```
//Translate doors
//Check if horizontal or vertical
lineHeight = | x(shortestPath[length of shortestPath – 1]) –
 x(shortestPath[length of shortestPath – 2]) |
lineWidth = | y(shortestPath[length of shortestPath – 1]) –
 y(shortestPath[length of shortestPath – 2]) |

if lineHeight>=lineWidth then
    for each door between shortestPath[length of shortestPath – 1]
in
rightDoors and targetRightDoors and leftDoors and
targetLeftDoors
and shortestPath[length of shortestPath – 2]
if y(shortestPath[length of shortestPath – 1]) <
y(shortestPath[length of shortestPath – 2]) then
        if door is bottom door then
                corrLeftDoor ++
            if targetDoor = door then
store in knowledge Base that target door is left door number
targetDoor
end if  end if
if door is top door then
            corrRightDoor ++
 if targetDoor = door then
store in knowledge Base that target door is right door number
targetDoor
 end if end if   end if
end for
end if
if lineHeight>=lineWidth then
    for each door between shortestPath[length of shortestPath – 1]
in
topDoors and targetTopDoors and bottomDoors and
targetBottomDoors
and shortestPath[length of shortestPath – 2]
        if y(shortestPath[length of shortestPath – 1]) <
y(shortestPath[length of shortestPath – 2]) then
        if door is right door then
                corrLeftDoor ++
            if targetDoor = door then
                store in knowledge Base that target
door is left door number targetDoor
 end if  end if
if door is left door then
            corrRightDoor ++
    if targetDoor = door then
            store in knowledge Base that target
door is right door number targetDoor
    end if  end if   end if
end for
end if
```

Algorithm 10: Translating Doors Algorithm Pseudo code

## IV. CONCLUSION

For the reason of the intelligent robot knows the directions of the optimal shortest path needed to reach its target in a new building work faster than intelligent robot that don't have any sort of external guidance where it needs to scan room by room and corridor by corridor to reach his target. From this point the need of such abilities have the high priority for mobile intelligent robot. Convert it to the knowledge base as a cognitive map inside the knowledge based system of the intelligent robot. In this research medial axis has been enhanced in order to be used for finding the optimal shortest path.